\documentclass[twoside,11pt]{article}

\usepackage{jmlr2e}
\usepackage{graphicx}
\usepackage{enumitem} 

\setlist[itemize]{noitemsep, topsep=1pt}

\ShortHeadings{Prediction of Kidney Function from Biopsy Images Using Convolutional Neural Networks}{Ledbetter, Lemley, MD and Ho}
\firstpageno{1}

\begin{document}

\title{Prediction of Kidney Function from Biopsy Images Using Convolutional Neural Networks}

\author{\name David Ledbetter \email dledbetter@chla.usc.edu \\
       \addr Children's Hospital Los Angeles\\
       Los Angeles, CA 
       \AND
       \name Long Van Ho \email loho@chla.usc.edu \\
       \addr Children's Hospital Los Angeles\\
       Los Angeles, CA 
       \AND
       \name Kevin V Lemley \email klemley@chla.usc.edu \\
       \addr Children's Hospital Los Angeles\\
       Los Angeles, CA }

\maketitle

\begin{abstract}
A Convolutional Neural Network was used to predict kidney function in patients with chronic kidney disease from high-resolution digital pathology scans of their kidney biopsies. Kidney biopsies were taken from participants of the NEPTUNE study, a longitudinal cohort study whose goal is to set up infrastructure for observing the evolution of 3 forms of idiopathic nephrotic syndrome, including developing predictors for progression of kidney disease. The knowledge of future kidney function is desirable as it can identify high-risk patients and influence treatment decisions, reducing the likelihood of irreversible kidney decline.
\end{abstract}

\section{Introduction}

\begin{figure}[htbp]
  \centering 
  \includegraphics[width=3.0in]{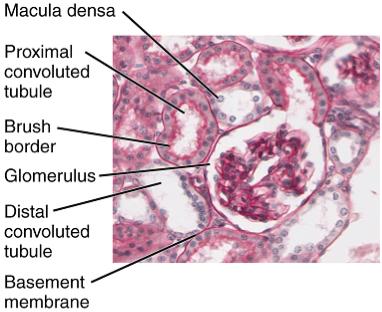} 
  \caption{Glomerulus and surrounding regions of the kidney \citep{wiki:glomerulus}.}
  \label{fig:glomerulus} 
\end{figure}

The measure of kidney function is estimated by how much primary filtrate from the blood passes through the glomeruli per minute, also known as the Glomerular Filtration Rate (GFR). The glomeruli (shown in Figure~\ref{fig:glomerulus}) serve as tiny filters that separate a watery filtrate from the rest of the cell and protein-containing components of the blood. The filtrate is then processed by the renal tubule to reclaim salts and nutrients and add metabolic wastes. In practice, however, exact GFR is difficult to measure and thus an approximation to the GFR is obtained based on serum creatinine measurements and demographic features. This quantity is known as the estimated Glomerular Filtration Rate (eGFR) and is commonly used as the indicator of kidney function and health. This paper discusses the development and training of the state-of-the art computer vision algorithm, the Convolutional Neural Network, to effectively predict eGFR 12 months in the future from baseline kidney biopsies.

\subsection{Motivation}

Previous work has correlated renal morphometry with changes in eGFR by hand-measuring properties such as the fractional interstitial area and average glomerular tuft volume \citep{lemley2008prediction}. Moreover, there are also obvious visual differences in kidney tissue between patients with stable and those with declining kidney function, further indicating the presence of an interaction between kidney function and glomerular form. In order to automate the extraction of visual information contained within the digital biopsy slides, a deep learning algorithm known as a Convolutional Neural Network (CNN) was trained to exploit the correlations between the morphometry in kidney biopsies and future kidney function. 

Convolutional Neural Networks have been utilized to great success in numerous vision classification problems including \citep{krizhevsky2012imagenet,szegedy2015going,he2015deep}. Additionally, CNNs have been utilized to extract information in other medical imaging tasks such as mitosis detection in breast cancer \citep{wang2014mitosis} and knee cartilage segmentation \citep{prasoon2013deep}.

\section{Data} 
This project utilized a subset of the NEPTUNE dataset -- a collaborative longitudinal study to research and set up infrastructure for observing predictors for idiopathic kidney disease \citep{gadegbeku2013design}. A subset of over 80 patients from this longitudinal study was available for processing. Initial biopsies of the participants were obtained at the beginning of the study and examined using Trichrome (TRI) and Periodic Acid-Schiff-diastase (PAS-D) slide staining techniques. Follow-up visits collected eGFR measurements at 4 to 6 month intervals for 5 years, which are used as targets for supervised training of the CNN. A kidney slide's resolution is on the orders of $(150000 \times 50000 \times 3)$ pixels with each pixel measuring $20$ microns. An example of a patient’s slide is shown below in Figure~\ref{fig:kidney_slide}.

\begin{figure}
 \centering
 \includegraphics[scale=.6]{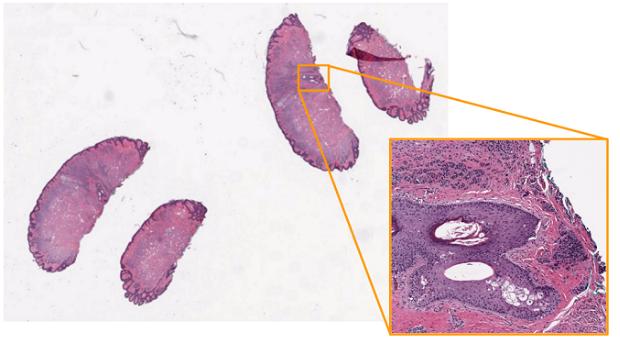} 
  \caption{Example of what a kidney biopsy looks like after processing.}
  \label{fig:kidney_slide} 
\end{figure} 

\subsection{Truth} 
Estimated Glomerular Filtration Rates (eGFR) were measured during the patients' first visit (baseline) and then at 4 month to 6 month intervals. The truth provided to the network was a single eGFR measurement a given time interval (e.g. at 12 month). Future work will incorporate a vectorized regression target to enable a broader range of clinically significant eGFR predictions (e.g. [4, 8, 12, 18, 24, 30, 36] month) to better reflect the overall trajectory of renal function.

\section{Preprocessing and Data Augmentation}

\subsection{Initial Try: Automated Segmentation}

\begin{figure}[htbp]
  \centering 
  \includegraphics[scale=.57]{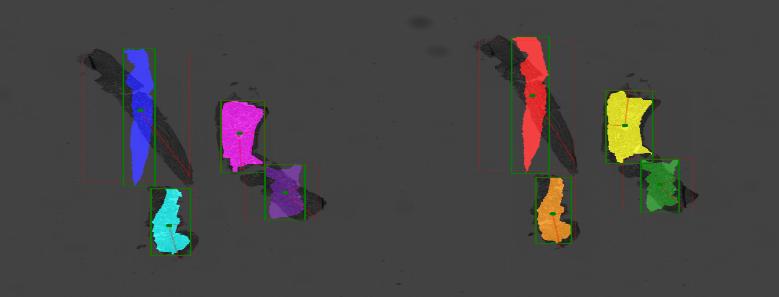} 
  \caption{Diagram depicting the segmentation and rotation of kidney sections to generate a minimum circumscribed bounding box.}
  \label{fig:kidney_segmentation} 
\end{figure}

Significant effort was dedicated to processing the data for inputs into the CNN framework. Initially, an automated segmentation algorithm was utilized to extract kidney segments from the complete biopsy slide. The algorithm used standard image processing and segmentation techniques such as histogram thresholding, erosion, and dilation to mask kidney tissue from noise and background. The segments were rotated along their major and minor axes to generate the minimum circumscribed bounding box. Results of the segmentation can be seen in Figure~\ref{fig:kidney_segmentation}. 

\begin{figure}[htbp]
  \centering 
  \includegraphics[scale=.6]{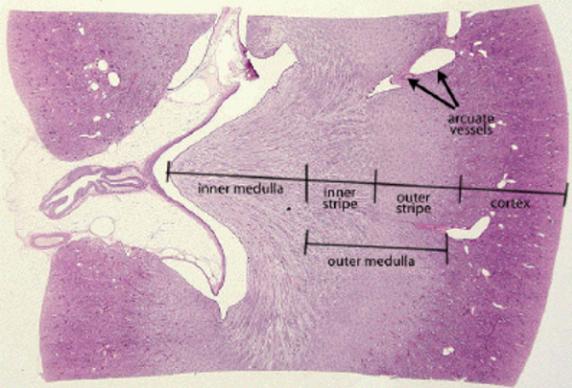} 
  \caption{Diagram depicting various regions of the kidney.}
  \label{fig:kidney_anatomy} 
\end{figure}

On one hand, the automated approach was effective at segmenting the kidney biopsy from the slide background; however, we did not discriminate between various portions of kidney biopsy. In particular, a significant fraction of kidney medulla was included in the automatically generated segments. Previous work \citep{lemley2008prediction} has indicated the cortex region of the kidney is more informative regarding kidney function. See Figure~\ref{fig:kidney_anatomy} for labeling of the kidney biopsy regions. As a result the segmented images generated via the automatic kidney segmentation algorithm were not utilized during training.

\subsection{Semi-Automatic Segmentation}

In order to focus on the primary goal of attempting to extract information from the kidney cortex, a semi-automatic algorithm was developed. Possible future work would include automatic further development of the automatic segmentation algorithm to more successfully mask the kidney cortex from the medulla.

The semi-supervised algorithm consisted of manually cropping the kidney cortex from the slide biopsies, referred to as Regions Of Interest (ROIs). This was quickly accomplished by utilizing the Leica ImageScope software (an interface capable of viewing, editing, and extracting ROIs from digital slides).  ImageScope was used to quickly generate ROIs of the kidney biopsies to contain mostly the kidney cortex and glomeruli. In clinical deployment, this would require a pathologist to manually extract ROIs from patient specimens prior to feeding it to the CNN predictive pipeline.

Using Leica's ImageScope software, a kidney database was generated containing segments of the kidney cortex over all the patients. There were on average 7 ROI extractions per slide, with resolutions ranging from $(2000 \times 2000 \times 3)$ to $(8000 \times 8000 \times 3)$ pixels.

After ROI extraction, 3 challenges remained to be addressed: 1) The data was still sparse, containing on average 35 ROI extractions per eGFR measurement; 2) ROI extraction resolution was much too large for practically training the CNN; and 3) ROI extractions had different resolutions and any common downsampling/upsampling to a common (height, width) would corrupt the physical shapes of the kidney tissue.

To address these challenges the ROIs were further processed into smaller image chips by cropping with a sliding window of size $(2000 \times 2000 \times 3)$, overlap of 50 percent and then downsampling by 2x. The resulting database 1) contained significantly more examples per patient; and 2) had manageable, uniform input resolutions $(1000 \times 1000 \times 3)$. An example of such an image chip can be seen below in Figure~\ref{fig:kidney_chip}.

\begin{figure}[htbp]
  \centering 
  \includegraphics[scale=.8]{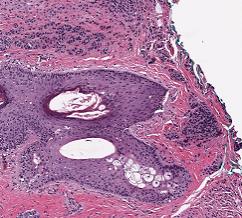} 
  \caption{Example of what a final kidney chip at $(1000 \times 1000 \times 3)$ resolution.}
  \label{fig:kidney_chip} 
\end{figure} 

In summary, the kidney biopsies collected from the NEPTUNE study were cropped to selected views of each patient’s kidney biopsy. These “image chips” would then be fed into the CNN for training, with each image chip paired with the patient’s 12 month eGFR. Finally, the predictions of each image chip per patients are averaged for the final eGFR prediction.

\subsection{Data Augmentation}

\begin{figure}[htbp]
  \centering 
  \includegraphics[scale=.7]{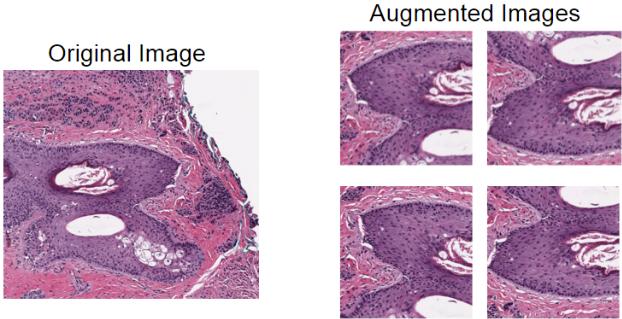} 
  \caption{Example random augmentations performed on incoming kidney biopsy chips used for training the Convolutional Neural Network.}
  \label{fig:chip_augmentation} 
\end{figure}

Upon loading the pre-processed database for training the CNN, the data is downsampled again by another 2x - 4x (resulting in images of size $(500 \times 500 \times 3)$ to $(250 \times 250 \times 3)$) and randomly augmented on-the-fly using the python package datumio \citep{datumio}. The following affine transformations were selected based on realistic expectations of the data:
\begin{itemize}
  \setlength{\itemsep}{1pt}
  \setlength{\topsep}{0pt}
  \setlength{\parskip}{1pt}
  \setlength{\parsep}{0pt}
  \item rotation: random angle between -15$^{\circ}$ and +15$^{\circ}$
  \item translation: random x,y translation of 7\%
  \item rescaling: random scale (zooming) factor of 5\%
  \item flipping: 50\% left/right and up/down symmetrical flipping
  \item cropping: after all previous augmentation, center crops of size $(400, 400, 3)$
\end{itemize}
The resulting inputs of the CNN were randomly perturbed views of the kidney biopsies of size $(400, 400, 3)$. Figure~\ref{fig:chip_augmentation} demonstrates an example of random augmentations applied to an example kidney view.

\section{Network Architecture}

The CNN architecture was heavily inspired by VGGNet \citep{simonyan2014very}, a very deep convnet that used small convolutional filters to construct deeper networks. A diagram of the complete network infrastructure can be found in Appendix A. Diverging from VGGNet was the injection of \textit{a priori} knowledge to the network. This was done by concatenating scaled vectors to the output of the second to last dense layer. This was called injection of ``aux-features" which included anything from hand-engineered features to the patient's age and sex. Inserting the features at the dense layer guides the network’s classification layers to not only leverage the learned compressed feature basis developed by the convolutions but also additional information extracted using \textit{a priori} knowledge to the network which is not available from the kidney biopsy images alone. Future works include implementing more recently successful techniques and layers such as Batch Normalization \citep{ioffe2015batch} and Residual Networks \citep{he2015deep}.

\subsection{Inputs}
Inputs to the network were image chips of each patient’s kidney slides and their associated aux-features. The images were passed to the convolution layers while the aux-features were appended to the second to last dense layer. Due to constraints of the data, the only aux-feature injected to network were baseline eGFR measurements (generally a strong predictor of subsequent eGFR). Future work will utilize hand-engineered features and patient attributes. The addition of the initial eGFR alone improved the network greatly -- decreasing the training time by 2x and the validation error of the network by $20\%$.

\section{Training} 

\subsection{Performance Metric \& Validation} 

For the network to be useful, it should be able to predict the eGFR of an entirely new patient. This means that the network should be transparent to never-before-seen biopsy slides, color-dyes, and digital imaging techniques. To properly evaluate performance based on these criteria, the training and validation sets were split based on labels of \textit{unique} patients. This enforces that any image chip (slice of kidney biopsy slide) associated with a patient cannot be included in both the training and validation set. This restriction aligns with our criterion in that the validation error represents the confidence of the network’s ability to extrapolate a never-before-seen patient’s eGFR 12 months into the future.
Moreover, due to the small number of truth constructs (a little over 80 unique patients, even less with adequate measurements and follow up data), a simple train/test split of 80/20 would leave the validation set with less than 16 patients. 5-fold patient-level cross-validations was used for more thorough investigation of network performance.
The primary metric used to evaluate model performance is a scatter plot of the true eGFR values (x) vs the predicted eGFR values (y). This was not a loss function used for optimizing the network, but served as an intuitive performance metric that is much easier to understand than a single number such as mean-squared-error. Qualitatively, the models can be compared based on how close the points are to a 1-to-1 line; quantitatively, the models can be compared based on the residuals of a least-squared linear fit to the predicted eGFR compared to the 1-to-1 line. 

\subsection{Optimizer and Hyper-Parameters} 

The models were trained using RMSProp \citep{dauphin2015rmsprop}, an adaptive learning rate that divides the current gradient by the moving average over the root-mean-squared of the weighted sum of the recent gradients. RMSProp can be seen as an extension of Adagrad \citep{duchi2011adaptive} with the addition of momentum. Hyper-parameters of RMSProp were left to their default values: $\rho=0.9$ and $\epsilon=1 \times 10^{-6}$ with an initial learning rate of $\mbox{lr}=0.0001$.
The learning rate was linearly decreased after every epoch (an entire loop through the training set). Weight updates were performed after every batch size of $32$. The network was trained utilizing a NVIDIA Titan X.
Future plans include investigations of other optimizers such as ADAM \citep{kingma2014adam} and ADADELTA \citep{zeiler2012adadelta}, hyper-parameter searches, and better learning rate procedures.

\subsection{Initialization} 

All layers of the network (with weights) were initialized using Glorot uniform (Xavier initialization) \citep{glorot2010understanding} which scales the weight elements to the number of parameters of input and the output of the layer. More specifically, each element of a layer’s weights draws from a uniform distribution with zero bias in the interval with $\mbox{W} = \mbox{U}\left(-\sqrt{6 / (n_{in} + n_{out})}, \sqrt{6 / (n_{in} + n_{out})}\right)$, where $n_{in}$ is the number of parameters feeding into the layer and $n_{out}$ is the number of output parameters of the layer.

\section{Results} 

\begin{figure}[htbp]
  \centering 
  \includegraphics[scale=.5]{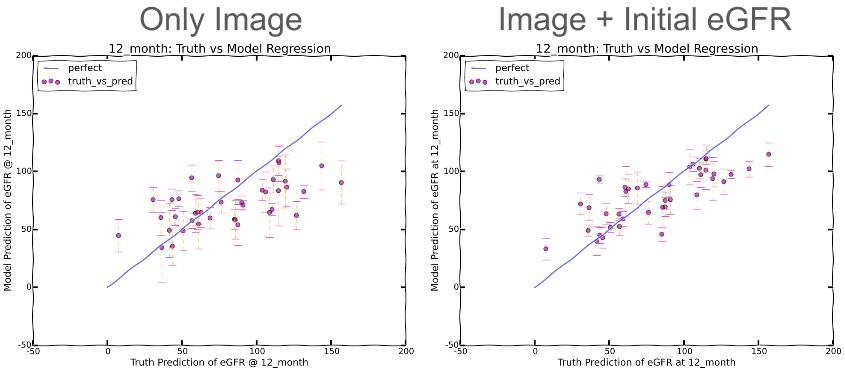} 
  \caption{Left: Truth vs. model predictions utilizing just image information. Right: Truth vs. model predictions incorporating initial eGFR information in the final dense layer of the CNN.}
  \label{fig:results_combined} 
\end{figure}

The preliminary network’s mean absolute error of predicting 12 months eGFR is $17.55$ ml/min. As a comparison, a simple propagation of the initial eGFR values to “predict” 12 months eGFR has an absolute error of $30.5$. This is a $42\%$ percent difference in model errors, illustrating that the network was able to learn useful features from patients’ kidney biopsies for predicting eGFR. The authors are currently in the process of obtaining 12 month eGFR predictions for all of the patients using another statistical method (generalized estimating equations) from another laboratory to compare performance on a standard benchmark. In Figure~\ref{fig:results_combined}, performance can be seen from the k-fold validation from two version of the network (left - no initial eGFR, right - with initial eGFR). Further work includes investigations with other CNN architectures and training techniques as well as unraveling the trained CNN to provide insights to the interactions between the kidney biopsies and future eGFR.

\section{Conclusions}

Several challenges were overcome including variety of laboratory standards in data collection, multiple staining techniques, image-scale which is not typically encountered in the image classification literature, and limited data availability (80 patients). Despite these challenges it was possible to extract visual information contained in the high-resolution digital pathology data for patients within the NEPTUNE study utilizing a Convolutional Neural Network. Several potential research opportunities remain moving forward including several pure machine learning improvements such as CNN architectures and hyper-parameter tuning as well as increased automation to augment the ability to perform analyses on additional data contained in the NEPTUNE dataset. However, initial results indicate the potential to continue to increase our ability to quickly and precisely extract relevant clinical predictions from high-resolution digital pathology products in order to improve our ability to provide clinicians with the information required to guide treatment strategies for patients suffering chronic kidney disease.


\bibliography{KidneyCHLA_MUCMD_2016}

\newpage
\appendix
\section*{Appendix A.}

\begin{figure}[htbp]
  \centering 
  \includegraphics[scale=.38]{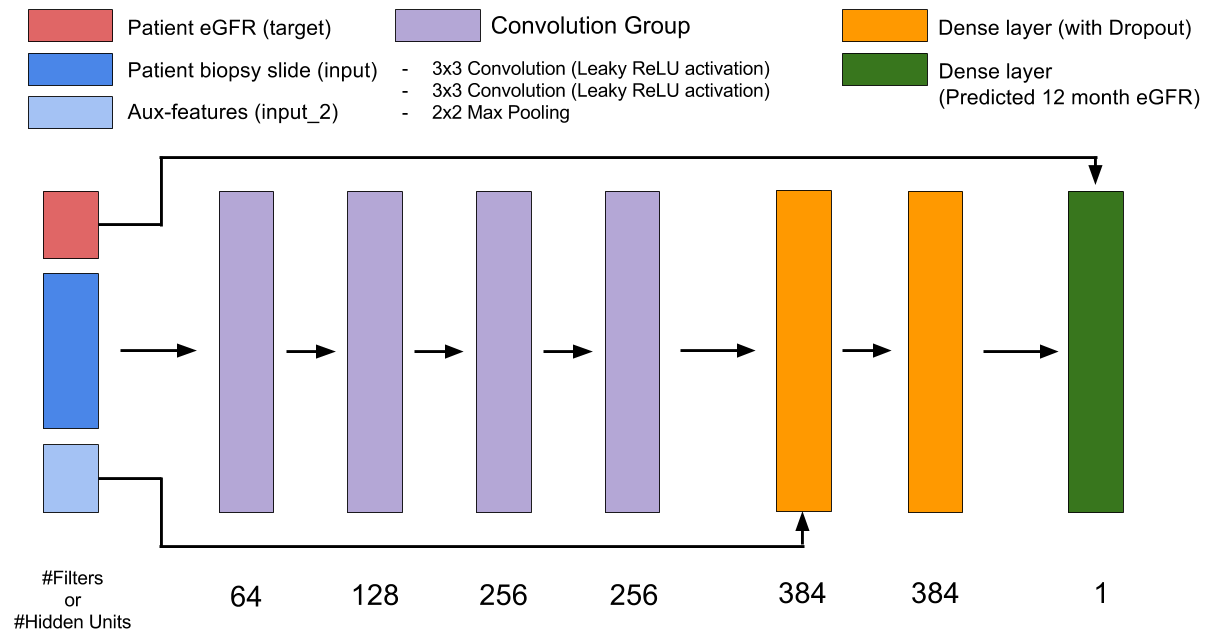} 
  \caption{Network infrastructure.}
  \label{fig:vgg16_kidneyv4_v3} 
\end{figure}

Figure \ref{fig:vgg16_kidneyv4_v3} illustrates the network architecture used to predict the future 12 month eGFR given kidney biopsies. Little work went into optimizing the hyper-parameters; for example, the number of Convolution Groups, filters within each group, and the number of hidden layer units in the dense layers. Future work will utilize more recently successful techniques such as Batch Normalization and Residual Networks, as well as further optimization of hyper-parameters.

\end{document}